\documentclass[conference]{IEEEtran}
\IEEEoverridecommandlockouts
\usepackage{cite}
\usepackage{amsmath,amssymb,amsfonts}
\usepackage{algcompatible}
\usepackage{algorithm}
\usepackage{algpseudocode}
\usepackage{graphicx}
\usepackage{textcomp}
\usepackage{xcolor}
\usepackage{caption} 
\def\BibTeX{{\rm B\kern-.05em{\sc i\kern-.025em b}\kern-.08em
    T\kern-.1667em\lower.7ex\hbox{E}\kern-.125emX}}
    
\makeatletter
\newcommand{\linebreakand}{%
  \end{@IEEEauthorhalign}
  \hfill\mbox{}\par
  \mbox{}\hfill\begin{@IEEEauthorhalign}
}
\makeatother

\begin{document}

\title{Physics-Informed Neural Networks  for Quantum Eigenvalue Problems}

%\author{\IEEEauthorblockN{Anonymous Author}
%\IEEEauthorblockA{\textit{Affiliation} \\
%\textit{Affiliation}\\
%Location \\
%Email}
%}

\author{\IEEEauthorblockN{Henry Jin}
\IEEEauthorblockA{\textit{School of Engineering and Applied Sciences} \\
\textit{Harvard University}\\
Cambridge, MA, USA \\
helinjin@g.harvard.edu}
\and
\IEEEauthorblockN{Marios Mattheakis}
\IEEEauthorblockA{\textit{School of Engineering and Applied Sciences} \\
\textit{Harvard University}\\
Cambridge, MA, USA \\
mariosmat@seas.harvard.edu}
\linebreakand
\IEEEauthorblockN{Pavlos Protopapas}
\IEEEauthorblockA{\textit{School of Engineering and Applied Sciences} \\
\textit{Harvard University}\\
Cambridge, MA, USA \\
pavlos@seas.harvard.edu}
}

\maketitle

\begin{abstract}
Eigenvalue problems are critical to several fields of science and engineering. We expand on the method of using unsupervised neural networks for discovering eigenfunctions and eigenvalues for differential eigenvalue problems. The obtained   solutions  are given in an analytical and  differentiable form that identically satisfies the desired boundary conditions. The network optimization is data-free and depends solely on the predictions of the neural network.  We introduce two physics-informed loss functions. The first, called \emph{ortho-loss}, motivates the network to discover pair-wise orthogonal eigenfunctions. The second loss term, called \emph{norm-loss}, requests the discovery of normalized eigenfunctions and is used to avoid trivial solutions. We find that embedding even or odd symmetries to the neural network architecture further improves the convergence for relevant problems. Lastly, a patience condition can be used to automatically recognize eigenfunction solutions. This proposed unsupervised learning method is used to solve the finite  well, multiple finite wells, and hydrogen atom eigenvalue quantum problems. 
\end{abstract}


\section{Introduction}
Differential equations are prevalent in every field of science and engineering, ranging from physics to economics. Thus, extensive research has been done on developing numerical methods for solving differential equations. With the unprecedented availability of computational power, neural networks hold promise in redefining how computational problems are solved or improving existing numerical methods.  Among other applications in scientific computing,  neural networks are capable of efficiently solving differential equations \cite{lagaris_eigen,nips2018,lagaris1998,mattheakis2020hamiltonian}.

These neural network solvers pose several advantages over numerical integrators:  the obtained solutions are analytical and differentiable \cite{lagaris1998}, numerical errors are not accumulated \cite{mattheakis2020hamiltonian}, networks are more robust against the ‘curse of dimensionality’ \cite{pnas2018, spiliopoulos2018},   a family of solutions corresponding to different initial or boundary conditions can be constructed \cite{cedric2020}, the neural solutions can be transferred for fast discovery of new solutions \cite{mattheakis2021unsupervised,desai2021oneshot}, inverse problems can be solved systematically \cite{Chen:20,paticchio2020semisupervised}, and available data can be incorporated into the loss function to improve the network's performance \cite{RAISSI2019686}.

Eigenvalue differential  equations with certain boundary conditions appear in a wide range of problems of applied mathematics and physics, including quantum mechanics and electromagnetism. Lagaris et al. \cite{lagaris_eigen} have shown that neural networks are able to solve eigenvalue problems and proposed a partially iterative method that solves a differential equation with a fixed eigenvalue at each iteration.
 More recently, Li et al. \cite{PhysRevA.103.032405} showed that neural networks can solve the stationary Schr\"odinger equation for systems of coupled quantum oscillators. This is a variational approach where the eigenvalue is indirectly calculated from the predicted eigenfunction.  
 Our work expands on the unsupervised neural network eigenvalue  solver presented by Jin et al. \cite{Jin2020UnsupervisedNN}, which simultaneously and directly learns the eigenvalues and the associated eigenfunctions using a scanning mechanism.   Here, we introduce physics-informed improvements to the regularization loss terms: orthogonal loss (ortho-loss) and normalization loss (norm-loss). We further design special neural network architectures with embedded symmetries that ensure the prediction of perfectly even or odd eigenfunctions.
 Furthermore, a modified parameterization is introduced to handle problems with  non-zero boundary conditions. The proposed technique is an extension to physics-informed neural network differential equation solvers and, consequently, inherits all the benefits that neural network solvers have over numerical integrators. 
 Moreover, our method has an additional advantage over integrators in that it discovers solutions that identically satisfy the boundary conditions. 
 We assess the performance of the proposed architecture by solving a number of standard eigenvalue problems of quantum mechanics: the single finite square well, multiple finite square wells, and the hydrogen atom.

\section{Background}

This study extends the method presented in \cite{Jin2020UnsupervisedNN}, where a fully connected neural network architecture was proposed, with a single output corresponding to the predicted eigenfunction, and with a constant input node designed to learn  constant eigenvalues through backpropagation. To identically satisfy the boundary conditions, a parametric function was used. 
In order for the network to find non-trivial solutions to the differential eigenvalue equation, the two regularization loss functions 
\begin{align}
\label{eq:oldNonTriv}
L_{f} = \frac{1}{f(x, \lambda)^2}, \quad    
    L_{\lambda} = \frac{1}{\lambda^2}
\end{align}
were used to penalize trivial eigenfunctions  and zero eigenvalues, respectively.
Moreover, a scanning mechanism allows the network to search the eigenvalue space for eigenfunctions of different eigenvalues, enabled by the loss term defined as  
\begin{align}
\label{eq:oldNonTriv}
L_{\text{drive}} = e^{-\lambda + c},
\end{align}
where $c$ was a value that changed during training through scheduled increases, and was used to control the scanning.

The research by Li et al. \cite{PhysRevA.103.032405} on neural network-based multi-state solvers is also relevant to this study. However, we present some novelties and differences in methodology. Specifically,  we assign a  trainable network parameter to discover the eigenvalue instead of indirectly calculating it through the expectation of the Hamiltonian of the system. Our approach avoids the repeated calculation of an integral (i.e., for the expectation value) which is evaluated every training epoch. The second novelty of our approach is the embedding of physical symmetries in the network architecture. The symmetry of the wavefunctions can be determined by the symmetry of the given potential function. We design a specialized architecture with embedded even or odd symmetry that significantly improves the overall network optimization. Finally, we suggest a parameterization that identically satisfies non-zero boundary conditions, which is necessary to solve the radial equation of the hydrogen atom.       % Specifically, we employ the use of a constant-input eigenvalue node, orthogonal loss, normalization loss, symmetry/antisymmetry, as well as non-zero parameterization.

Orthogonality loss is also used in \cite{PhysRevA.103.032405}, where it is leveraged to simultaneously produce multiple eigenvalue solution outputs that are pair-wise orthogonal. This differs from our method, since our neural network outputs one solution at a time, and the orthogonality loss term is used to prevent us from finding the same solution multiple times.
%%%%%%%%%%%%%%%%%%%%%%%%%%%%%%%%%%%%%%%%%%%%%%%%%%%%%%%%%%%%%%%%%%%%%%%%%%%%%%%%%%%%%%%%
\section{Methodology}

\label{gen_inst}

We consider an eigenvalue problem that exhibits the form:
\begin{align}
\label{eq:EigenProblem}
\mathcal{L}f(x) = \lambda f(x),
\end{align}
where $x$ is the spatial variable, $\mathcal{L}$ is a differential operator that depends on $x$ and its derivatives, $f(x)$ is the eigenfunction, and $\lambda$ is the associated eigenvalue. For the finite square well problems, we assume homogeneous Dirichlet boundary conditions at the left and right boundaries $x_{L}$ and $x_{R}$, respectively, such that $f(x_L)=f(x_R)=f_b$, where $f_b$ is a given constant boundary value. On the other hand, for the hydrogen atom problem, a single Dirichlet boundary condition $f(x_R)=f_b$ is enforced.

\begin{figure}[h]
    \centering
    \includegraphics[width = 0.5\textwidth]{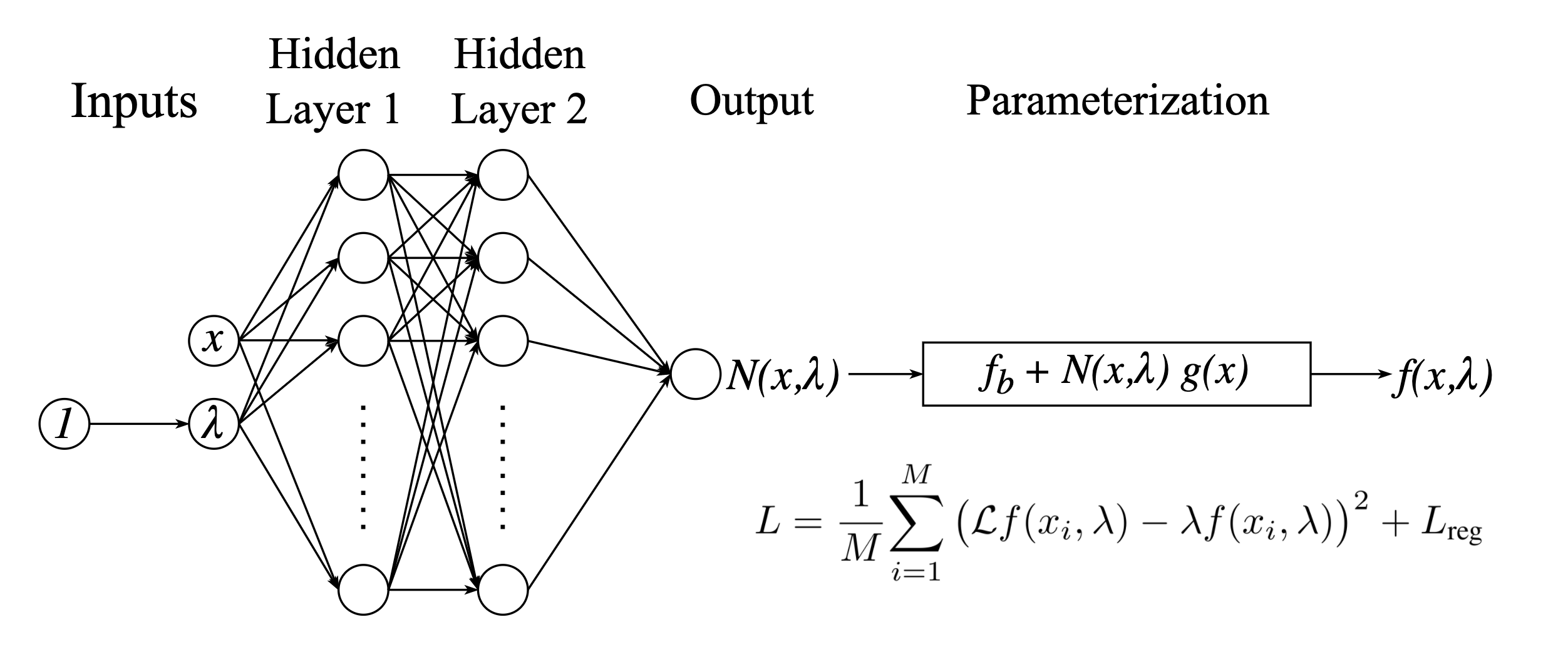}
    \caption{Physics-informed neural  architecture for solving   eigenvalue problems. }
    \label{fig:finalarch}
\end{figure}

We expand on the network architecture proposed by  \cite{Jin2020UnsupervisedNN} and shown in Fig. \ref{fig:finalarch}.
 This feed-forward neural network  is capable of solving Eq. (\ref{eq:EigenProblem}) when both $f(x)$ and $\lambda$ are unknown. The network takes two inputs, the variable $x$ and a constant input of  ones. The constant input  feeds into a single linear neuron (affine transformation) that is updated through optimization, allowing the network to find a constant $\lambda$. Afterwards, $x$ and $\lambda$ are inputs to a fully-connected feed-forward neural network that returns an output function $N(x,\lambda)$. The predicted eigenfunctions $f(x,\lambda)$ are defined using a parametric trick, similar to  \cite{mattheakis2020hamiltonian}, according to the equation:
\begin{align}
\label{eq:parSol}
    f(x,\lambda) = f_b + g(x)N(x,\lambda).
\end{align}
By choosing an appropriate $g(x)$, the predicted eigenfunction identically satisfies certain  boundary conditions.

Our  aim   is to discover pairs of $f(x, \lambda )$ and $\lambda$  that approximately satisfy Eq. (\ref{eq:EigenProblem}). This is achieved by minimizing, during the network optimization, a loss function $L$   defined by Eq. (\ref{eq:EigenProblem}) as:
\begin{align}
\label{eq:Loss}
    L &=  L_\text{DE} +  L_\text{reg} \nonumber \\
    L  &= \frac{1}{M} {\sum_{i = 1}^{M} \big( \mathcal{L}f(x_i, \lambda) - \lambda f(x_i, \lambda) \big)^2} +  L_\text{reg},
\end{align}
where averaging  with respect to $x_i$ takes place in $L_\text{DE}$ for $M$ training sample points, namely $x=(x_1,\cdots,x_M)$. Any derivative with respect to $x_i$ contained in  $\mathcal{L}$ is calculated by using the auto-differentiation technique \cite{Paszke2017AutomaticDI}.
The $L_\text{reg}$ term in Eq. (\ref{eq:Loss}) contains regularization loss terms. In this work, we introduce and apply a regularization function that consists of three terms of the form: $L_\text{reg} = \nu_\text{norm} L_\text{norm} +\nu_\text{orth} L_\text{orth} + \nu_\text{drive} L_\text{drive}$. { Empirically, for the problems discussed below, we found  the optimal regularization coefficients   $\nu_\text{norm}= \nu_\text{orth}=1$. } The normalization loss $L_\text{norm}$ encourages normalized eigenfunctions, avoiding the discovery of trivial eigenfunctions and eigenvalues, since  it  enforces non-zero solution as well as constraining the eigenfunction's squared integral to be finite. The $ L_\text{orth}$  motivates the network to scan for orthogonal eigenfunctions and can  replace or assist the non-physical scanning ($L_\text{drive}$) method used in \cite{Jin2020UnsupervisedNN}. $L_\text{drive}$ accounts to the scanning method which is  used to guide the model's eigenvalue weight and is given by Eq. \ref{eq:oldNonTriv}. However, for the experiments presented in this study, we use $L_\text{orth}$ as a physics-informed regularization term that can replace the non-physical scanning method with $L_\text{drive}$, and thus $\nu_\text{norm}$ is set to 0. 
%%%%%%%%%%%%%%%%%%%%%%%%%%%%%%%%%%%%%%%%%%%%%%%%%%%%%%%%%%%%%%%%%%%%%%%%%%%%%%%%%%%%%%%%
\subsection{Normalization Loss}

Our contribution includes a novel approach to solving the trivial solution problem. While \cite{Jin2020UnsupervisedNN} employed non-trivial eigenfunction and non-trivial eigenvalue loss terms $L_f$ and $L_\lambda$, as described in Eq. \ref{eq:oldNonTriv}, these loss terms cannot numerically converge to 0 without scaling the solutions to infinity, and thus they introduce numerical error. While they were effective for preventing the network from converging to trivial $f(x)$ and $\lambda$, they hold no physical meaning. We present a physics-aware regularization loss function that not only prevents trivial solutions, but also motivates the eigenfunction's inner product with itself to approach a specific constant number, which is the normalization constraint physically required of eigenfunctions in quantum mechanics. Thus, $L_\text{norm}$ is given by

\begin{align}
\label{eq:normloss}
    L_{\text{norm}} = \left(f(x,\lambda) \cdot f(x,\lambda) - \frac{M}{x_R-x_L}\right)^2,
\end{align}
where dot denotes the inner  product. The loss function in Eq. (\ref{eq:normloss}) drives the network to find solutions with non-zero integrals, where $f(x,\lambda)$ represents the network solution, $M$ is the number of samples, and $x_R-x_L$ is the training range. Specifically, this motivates the network solution to have a squared integral equal to one. Unlike $L_f$ and $L_\lambda$, $L_\text{norm}$ can strictly reach zero and can also satisfy the normality constraint for eigenfunction solutions of Schrodinger's equation.

%%%%%%%%%%%%%%%%%%%%%%%%%%%%%%%%%%%%%%%%%%%%%%%%%%%%%%%%%%%%%%%%%%%%%%%%%%%%%%%%%%%%%%%%
\subsection{Orthogonality Loss}
An orthogonality loss regularization function  is included as part of $L_\text{reg}$ to motivate the network to find different eigen-solutions to  Schrodinger equation. This presents a physics-informed approach whereby we can motivate a network to solve for orthogonal solutions for problems where it is known that solutions are orthogonal, a fundamental property of linear differential eigenvalue problems. Schrodinger's equation is one such example, but this mechanism can be extended to any Hermitian operator. This serves as a replacement or an improvement over solely relying on the scanning mechanism $L_\text{drive}$ presented  in \cite{Jin2020UnsupervisedNN}. While a scanning search through the eigenvalue space using $L_\text{drive}$ can be useful for providing control over the model's search for eigenfunction solutions, solving equations that are known to be Hermitian (such as the Schrodinger equation) allows the use of an orthogonal loss term, since eigenfunctions of Hermitian operators are orthogonal. In this paper, we show that the neural network is able to find orthogonal eigenfunction solutions solely based on the orthogonality loss. This loss term is given by the following equation.

\begin{align}
\label{eq:ortholoss}
    L_{\text{orth}} = \psi_{\text{eigen}} \cdot \psi,
\end{align}
where $\psi_{\text{eigen}}$ denotes the sum of all eigenfunctions that have already been discovered by the network during training, and $\psi$ is the current network prediction. This regularization term embeds the network with a physics-informed predisposition towards finding orthogonal solutions to a Hermitian operator, serving as a more physics-aware loss term than the brute-force scanning approach.

Following the network's convergence to a new solution, the new eigenfunction is added to $\psi_{\text{eigen}}$ and  thus, it is the linear combination of all the discovered solutions. Hence,  a single orthogonality loss term is computed for each learning gradient, as opposed to separate orthogonality computations for each learned eigenfunction. This reduces computational cost since only one dot product is computed for each training iteration, as opposed to multiple dot products with each found eigenfunction.

%%%%%%%%%%%%%%%%%%%%%%%%%%%%%%%%%%%%%%%%%%%%%%%%%%%%%%%%%%%%%%%%%%%%%%%%%%%%%%%%%%%%%%%%
\subsection{Embedding Even and Odd Symmetry}
For certain differential equations where prior information about the potential dictates even or odd symmetric  eigenfunctions, the neural network architecture can be embedded with a physics-informed modification that enforces the correct symmetry in the eigenfunction output. As demonstrated by Mattheakis et al. in \cite{mattheakis2020physical} and extended by \cite{DBLP:journals/corr/abs-2106-12891}, symmetry can be embedded by feeding a negated input stream in parallel to the original input, then combining streams before the final dense layer. Adding streams leads to even symmetric outputs, while subtracting gives rise to odd symmetric predictions.

We  found that embedding symmetry into our model significantly accelerates the convergence to a solution. This is relevant for the multiple finite square wells problem, as we demonstrate below. 

%%%%%%%%%%%%%%%%%%%%%%%%%%%%%%%%%%%%%%%%%%%%%%%%%%%%%%%%%%%%%%%%%%%%%%%%%%%%%%%%%%%%%%%%
\subsection{Parametric Function}
Selecting an appropriate parametric function $g(x)$ is necessary for enforcing boundary conditions. The following parametric equation enforces a $f(x_L)=f(x_R)=0$ boundary conditions:
\begin{align}
\label{eq:parametricDouble}
    g(x) = \left(1-e^{-(x-x_L)}\right)\left(1-e^{-(x-x_R)}\right).
\end{align}

As demonstrated in \cite{Jin2020UnsupervisedNN}, this parametric function is suitable for problems  where the eigenfunctions are fixed to or converge to zero, as in the case of the infinite square well and the harmonic oscillator problems. In the following experiments,  we employ the  parametric function of Eq. (\ref{eq:parametricDouble}) for finite square well problems, as they similarly require eigenfunctions to taper to zero at domain limits.

The differential eigenvalue equation for the hydrogen atom, however, has a single zero boundary condition at $x\rightarrow \infty$, as the fundamental solution is not fixed to 0 at the origin. For such problems where a single Dirichlet boundary condition is required, we use the following parametric function:
\begin{align}
\label{eq:parametricSingle}
    g(x) = \left(1-e^{-(x-x_R)}\right).
\end{align}

%%%%%%%%%%%%%%%%%%%%%%%%%%%%%%%%%%%%%%%%%%%%%%%%%%%%%%%%%%%%%%%%%%%%%%%%%%%%%%%%%%%%%%%%
\subsection{Towards Solution Recognition}
To automatically extract the correct eigenfunctions, we define convergence to an eigenfunction solution using two criteria: the differential equation loss $L_\text{DE}$ and patience.

$L_\text{DE}$ describes the loss term for the differential eigenvalue equation in question. For our experiments, without loss of generality, we used  Schrodinger's equation. Nevertheless, the method is valid for any  differential equation  eigenvalue problem. Considering that perfect eigenvalue solutions will have an $L_\text{DE}$ loss equal to zero, we claim that a solution is found when $L_\text{DE}$ falls below a chosen threshold, which is a hyper-parameter in the training process.

The patience condition describes the model's training progress. When solving for a solution, the model initially improves very quickly, resulting in a fast decrease of $L_\text{DE}$. However, over the course of converging to a solution, the rate of decrease in $L_\text{DE}$ decreases as well. Thus, we use the rate of decrease in $L_\text{DE}$ as another condition for solution recognition. If the rolling average during the training iterations of the successive differences in $L_\text{DE}$ over a specified window hyper-parameter falls below a chosen threshold hyper-parameter, we consider the patience condition to be met.

When both the $L_\text{DE}$ condition ($L_\text{DE}$ falling below a threshold) and the patience condition are satisfied, we consider an eigenvalue solution to have been  found. On the other hand, if only the patience condition is satisfied, then we interpret this to mean that the model has converged to a false solution. Consequently, we switch the symmetry (from even to odd symmetry or vice versa) of the model to motivate the network to search for other solutions. This approach of switching the symmetry of the model upon converging to a false solution was inspired by our finding that the network's function output after converging to false solutions resembled true solutions, but of the opposite symmetry. Upon adopting this switching approach, we found that the model was able to resume finding true solutions. The above method is described by Algorithm  \ref{alg:cap}.

\begin{algorithm}
\caption{The Physics-Informed Neural Eigenvalue Solver Algorithm}\label{alg:cap}
\begin{algorithmic}[1]
\State Instantiate model with even symmetry
\While{training}
\State Generate training samples $x_i$
\State Compute $L_\text{DE}$, $L_\text{norm}$
\State Compute $L_\text{orth}$ using all stored eigenfunctions
\State Backpropagate and step
\If{patience condition  {\bf and} $L_\text{DE}<\text{threshold}$}
    \State Store copy of model
\ElsIf{patience condition}
    \State Switch model symmetry
\EndIf
\EndWhile
\end{algorithmic}
\end{algorithm}

%%%%%%%%%%%%%%%%%%%%%%%%%%%%%%%%%%%%%%%%%%%%%%%%%%%%%%%%%%%%%%%%%%%%%%%%%%%%%%%%%%%%%%%%
\section{Experiments}
We evaluate the effectiveness of the proposed method  by solving eigenvalue problems defined by Schrodinger's equation. Schrodinger's equation is the fundamental equation in quantum mechanics that  describes the state wavefunction $\psi(x)$  and the associated energy $E$ of a quantum system. In this study, we are interested in solving the one-dimensional stationary  Schrodinger's equation  defined as:
\begin{align}
\label{eq:QMeigen}
    % {H}\psi(x) &= E\psi(x) \nonumber \\
    \left[ -\frac{\hbar^2}{2m}\frac{\partial^2}{\partial x^2} + V(x)\right] \psi(x) = E\psi(x),
\end{align}
where $\hbar$ and $m$ stand for the reduced Planck constant and the mass respectively, {which} without loss of  generality, can be set to   $\hbar=m=1$.   Equation (\ref{eq:QMeigen}) defines an eigenvalue problem where $\psi(x)$ and $E$ denote the {eigenfunction $f(x,\lambda)$ and eigenvalue $\lambda$ pair}. The differential equation loss for this one-dimensional stationary Schrodinger's equation is given by Equation (\ref{eq:LSE}), and henceforth we call this the Schrodinger equation loss.

\begin{align}
\label{eq:LSE}
    L_\text{DE} &= \frac{1}{M} {\sum_{i = 1}^{M} \big( \left[ -\frac{\hbar^2}{2m}\frac{\partial^2}{\partial x^2} + V(x_i)\right]f(x_i, E) - E f(x_i, E) \big)^2}.
\end{align}

A  boundary condition eigenvalue problem is defined by considering a certain potential function $V(x)$ and boundary conditions for $\psi(x)$.
We assess the performance of the proposed network architecture by solving Eq. (\ref{eq:QMeigen}) for the potential functions of the single finite  well, multiple coupled finite  wells, and the radial equation for the hydrogen atom, all of which have known analytical solutions.

For the training, a batch of $x_i$ points in the interval $[ x_L, x_R ] $ is selected as input. In every training iteration (epoch) the input points are perturbed by a Gaussian noise to prevent the network from learning the solutions only at fixed points. Adam optimizer is used with a learning rate of $8 \cdot 10^{-3}$.
%
% \cite{mattheakis2020hamiltonian},  \cite{adam}
We use two hidden layers of 50 neurons per layer with  trigonometric $\sin(\cdot)$  activation function.  The use of  $\sin(\cdot)$ instead of more common activation functions, such as Sigmoid$(\cdot)$ and $\tanh(\cdot)$, significantly accelerates the network's convergence to a solution \cite{mattheakis2020hamiltonian}.  We  implemented  the proposed neural network in pytorch  \cite{Paszke2017AutomaticDI} and published the code on github\footnote{https://github.com/henry1jin/quantumNN}.

% 

%%%%%%%%%%%%%%%%%%%%%%%%%%%%%%%%%%%%%%%%%%%%%%%%%%%%%%%%%%%%%%%%%%%%%%%%%%%%%%%%%%%%%%%%
\subsection{Single Finite  Well}

The finite  well potential function is defined  as:
%similarly as the infinite square well, but with finite constant potential: 
\begin{equation}
\label{eq:finpot}
V(x) = \begin{cases} 
      %0 & -\frac{\ell}{2}\leq x\leq \frac{\ell}{2}, \\
      0 & 0\leq x\leq \ell \\
      V_0 & \text{otherwise} 
   \end{cases},
\end{equation}
where $\ell$ is the length and $V_0$ is the depth  of the quantum well. 
The analytical solution to the finite  well problem is traditionally found by solving the stationary Schrodinger's equation in each region, then 'stitching' the solutions of each region together while enforcing a continuous eigenfunction that is also continuously differentiable. For bound eigenfunctions, the general form of the solution for regions where the eigenvalue $E$ is greater than the potential reads: 
\begin{equation}
\label{eq:finsolslower}
\psi = A\sin(kx) + B\cos(kx), \quad
k = \frac{\sqrt{2mE}}{\hbar}.
\end{equation}
For regions where the eigenvalue $E$ is smaller than the potential energy, the solution's general form is

\begin{equation}
\label{eq:finsolshigher}
\psi = Ce^{-\alpha x} + De^{\alpha x}, \quad
\alpha = \frac{\sqrt{2m(V_0-E)}}{\hbar}.
\end{equation}

The solutions then for Eq. (\ref{eq:finpot}) is the following piece-wise eigenfunction, where constants $c_1$,  $c_2$, and $\delta$, are determined by the requirement that the eigenfunction is continuous, continuously differentiable, and normalized. 

\begin{equation}
\label{eq:finsolexample}
\psi(x) = \begin{cases} 
      c_1 e^{\alpha x} & x\leq 0, \\
      c_2 \sin(kx+\delta) & 0 < x\leq \ell, \\
      c_1 e^{-\alpha x} & x > \ell
   \end{cases}
\end{equation}
The $\psi(x)$ eigenfunctions must decay to infinity outside the walls, implying the boundary conditions $ \psi(-\infty)= \psi(\infty)=0$. In numerical methods, infinity is approximated with large values relative to the potentials. We adopt the approximate boundary conditions of $\psi(x_L) = \psi(x_R) = 0$ with the choice $x_L=x_R=6\ell$,  for $\ell=1$ and   $V_0=20$. The proposed model with the orthogonal loss term is capable of solving for all bound eigenstates. In the following we use the neural network to approximate the first four eigenfunctions and the associated energies.

\begin{figure}[h]
    \centering
    \includegraphics[scale=.34]{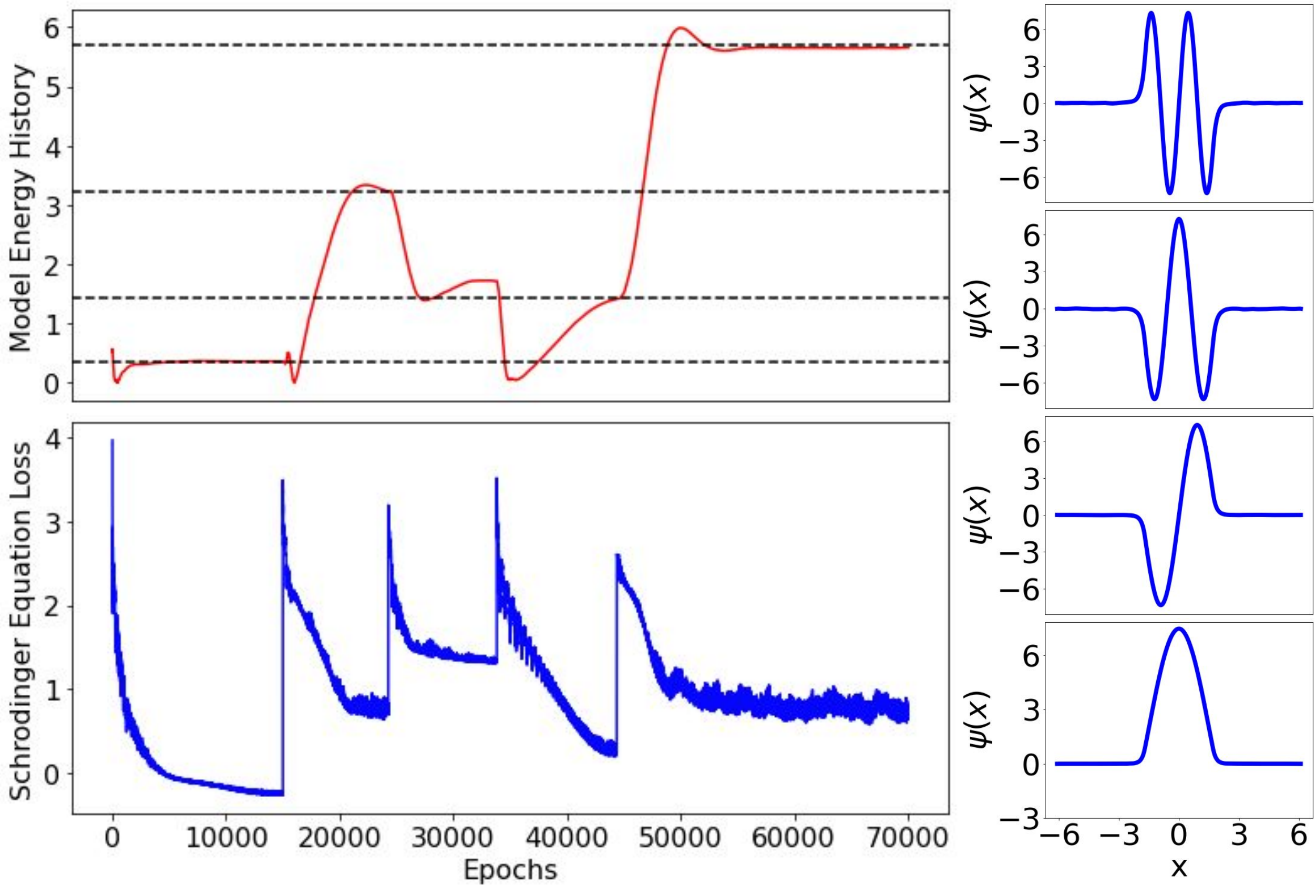}
    \caption{The plot on the upper left displays the model's eigenvalue weight (i.e. energy) during the training process. Horizontal dashed black lines demarcate the true eigenvalues, which our network accurately finds (plateaus of red lines). The lower left plot shows the corresponding Schrodinger equation loss $L_\text{DE}$ over epochs during training. At 15000 epochs, orthogonal loss with the first eigenfunction is introduced. Each following spike in  $L_\text{DE}$ indicates the point where the model reaches the patience condition, and the last eigenfunction is added to the orthogonal loss term. Column of plots on the right are the resulting eigenfunctions that the model finds when the predicted energy converges to a plateau.}
    \label{fig:singlefinite}
\end{figure}

We start the network optimization by using a neural network with even symmetry embedded. Figure \ref{fig:singlefinite} summarizes the results for the discovery of the first four eigenstates of the quantum finite well. The lower left panel outlines the $L_\text{DE}$  during the training. The red curve in the upper left graph demonstrates the predicted energies where the plateaus indicate the  discovery of an eigenstate; the dashed black lines show the ground truth energies. On the right side, the four predicted eigenfunctions are  represented by blue lines;  the bottom graph corresponds to the  ground state. In particular, the neural network finds for the ground state solution with energy $E = 0.3586$. After the first solution is found, we introduce the orthogonal loss term into the training, motivating the network to find a new eigenfunction. Consequently, the eigenvalue weight departs from its first value and rises to find the next even-symmetry solution with eigenvalue $E = 3.2132$ (the third graph on the right side in Fig. \ref{fig:singlefinite}, counting from the bottom). Once the patience condition is reached, the network automatically adds the latest solution to the orthogonal loss, motivating the network to once again depart its solution in search of the next orthogonal solution. The model converges to an eigenvalue of around $E = 1.8$, however it does not meet both conditions for solution acceptance. In particular, it does not meet the $L_\text{DE}$ condition. We take this to mean that, while the model has converged, it has converged to a false solution. So the symmetry of the model is switched to odd symmetry. The next two solutions found are odd-symmetric and correspond to the eigenvalues of $E = 1.4322$ and $E = 5.6873$ shown respectively by the second and fourth images in the right panel of  Fig. \ref{fig:singlefinite}.

%%%%%%%%%%%%%%%%%%%%%%%%%%%%%%%%%%%%%%%%%%%%%%%%%%%%%%%%%%%%%%%%%%%%%%%%%%%%%%%%%%%%%%%%
\subsection{Multiple Finite Square Wells}
The single finite  well potential can be repeatedly spaced to create a potential function that consists of multiple square wells as follows:

\begin{equation}
\label{eq:multiplefinpot}
V(x) = \begin{cases} 
      0 & 2n\ell\leq x\leq (2n+1)\ell \\
      V_0 & \text{otherwise}
   \end{cases},
\end{equation}
where $ n $ is an element of a subset of nonzero integers.

Like the single finite  well, solutions to the multiple square wells are piece-wise constructed by solving for each discrete region and "stitching" solutions. The general forms of the solutions in each region, namely, inside and outside a well,  are once again given by Eq. (\ref{eq:finsolslower}) and Eq. (\ref{eq:finsolshigher}), respectively. The boundary conditions at infinity are also approximated by large values of $x$ relative to the potential, that is, $\psi(x\rightarrow \pm \infty)=0$.

\begin{figure}[h]
    \centering
    \includegraphics[scale=.43]{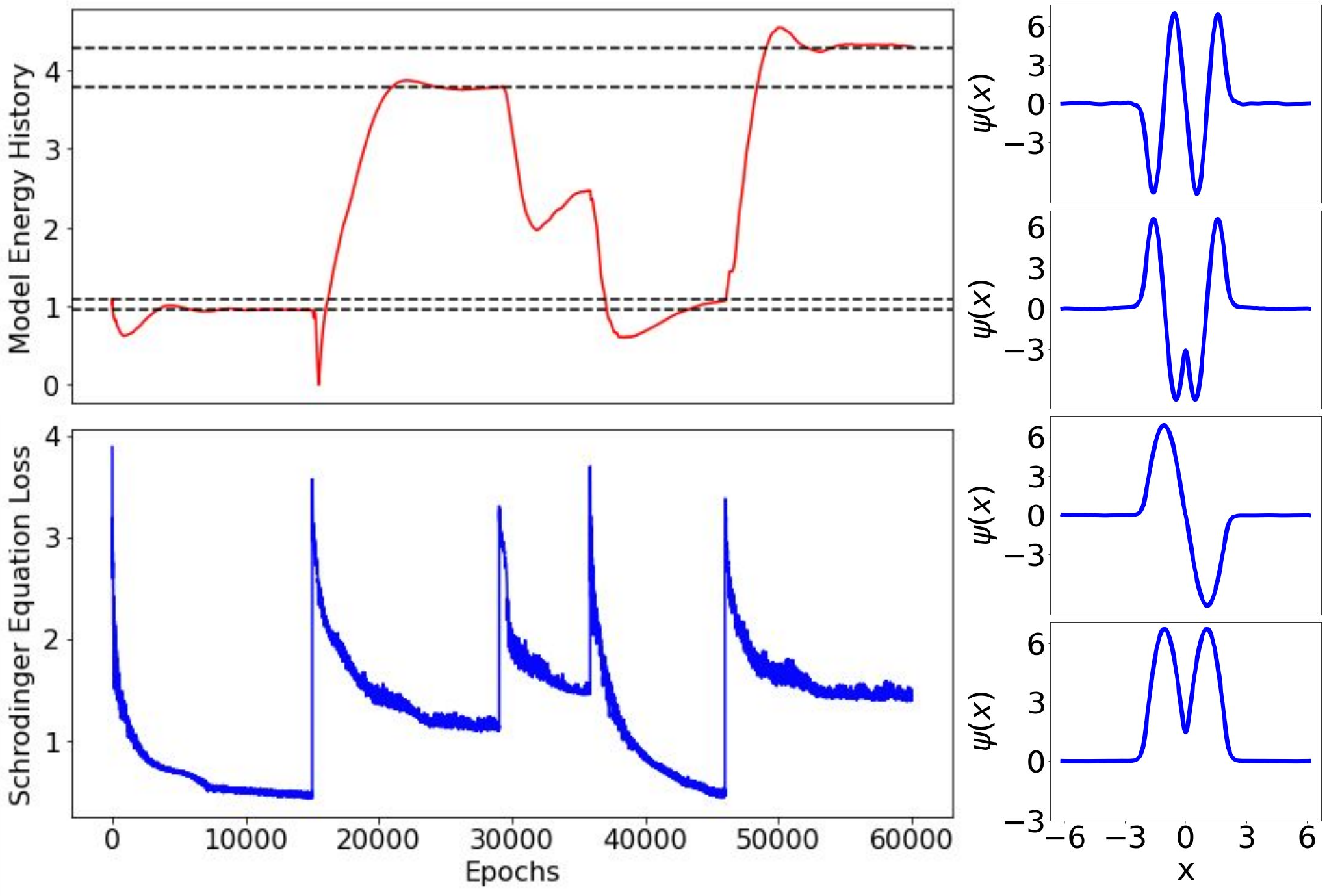}
    \caption{Top left plot shows the network's eigenvalue over epochs during training, with true eigenvalues shown by the dotted horizontal lines. Bottom left plot shows the model's corresponding Schrodinger equation loss at each training point. Column of plots on the right are the resulting eigenfunctions that the model finds. }
    \label{fig:multiplefinite}
\end{figure}

Our deep learning technique applied to the multiple  wells solves for an arbitrary number of the solutions. Figure \ref{fig:multiplefinite} shows our neural network finding the four lowest-energy (i.e. lowest-eigenvalue) states of the double finite square well. Similar to the single finite  well problem, the model here uses the physics-informed approach of solving for orthogonal eigenfunctions with the orthogonal loss term, given the knowledge that solutions to Hermitian differential operators must be orthogonal.

\subsection{Symmetry vs No Symmetry}

\begin{figure}[h]
    \centering
    \includegraphics[scale=.35]{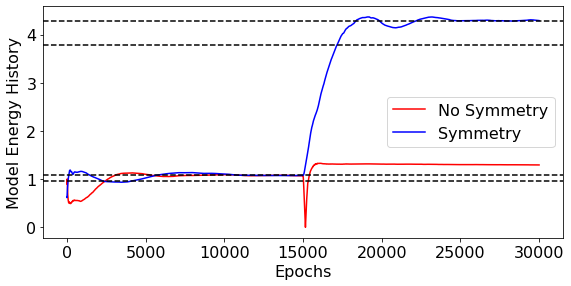}
    \caption{Predicted eigenvalue energies of quantum multiple wells during the training of a network with embedded symmetry (blue) and a network without any embedded symmetry (red).  
 Dotted black horizontal lines are analytical, ground truth solution eigenvalues. }
    \label{fig:symm_no_symm}
\end{figure}

Embedding symmetry into the network for problems where the solutions are known to be either even or odd symmetric proved to greatly improve the solution accuracy. Figure \ref{fig:symm_no_symm} compares  the eigenvalues (energy) predicted by the two models, one with embedded symmetry (blue line) and one without (red line). While the symmetry-embedded model is able to smoothly transition from one correct eigenvalue to the next one, the model without embedded symmetry converges to an incorrect eigenvalue.

%%%%%%%%%%%%%%%%%%%%%%%%%%%%%%%%%%%%%%%%%%%%%%%%%%%%%%%%%%%%%%%%%%%%%%%%%%%%%%%%%%%%%%%%
\subsection{Hydrogen Atom}
In quantum mechanics, the  hydrogen atom is described by the three-dimensional Schrodinger equation with a Coulomb potential energy. While the hydrogen atom is a three-dimensional  problem, the equation can be decomposed into radial and angular components via separation of variables. The angular equation yields the spherical harmonics solutions, while the radial component $R(r)$ equation to be solved is  given by:
\begin{equation}
\label{eq:radialHydrogen}
\frac{d^2R}{dr^2} + \frac{2}{r}\frac{dR}{dr} = - \left( \frac{2\mu}{\hbar^2}\left(E+\frac{Ze^2}{4\pi\epsilon_0 r}\right)-\frac{l(l+1)}{r} \right) R,
\end{equation}
where $r$ is the radial variable, $\hbar$ is the reduced Planck's constant, $\mu$ is the reduced mass,  $Z$ is the number of protons, $\epsilon_0$ denotes the vacuum permeability, and the variable $l$ denotes the angular momentum of the system and takes positive integer values. We employ the proposed neural network to solve Eq. (\ref{eq:radialHydrogen})  for $l = 0,1,2,3$. 

We note that Eq. \ref{eq:radialHydrogen} becomes singular at $r = 0$. Consequently, training sample points close to $r = 0$ lead to numerical instability. To avoid this problem, we allocate the region $r = [0, 1e-1]$ to be a no-train zone. Thus, any training sample points that are generated are greater than $r = 1e-1$. Without this constraint, the numerical instability caused by sample points close to 0 disrupts the network's ability to converge to solutions.

The analytical eigenvalue energies of the hydrogen atom are given by 
\begin{equation}
\label{eq:H_E}
E_n = -\frac{\mu Z^2 e^4}{32 \epsilon_0^2 \hbar^2 \pi^2 n^2},
\end{equation}
where $n$ denotes the order of the  solutions. Namely, for $n=1$, we get the ground energy. We notice that the eigenvalue energies are not dependent on the system's angular momentum $l$, but only on the system's order of excitement $n$.

The full three-dimensional solution to the Schrodinger equation for the hydrogen atom creates probability densities. The densities have not only radial dependence, but also angular dependence. For our work, we focused solely on the radial component of the Schrodinger equation. Furthermore, without loss of generality, we set $\frac{\mu Z^2 e^4}{8 \epsilon_0^2 h^2 n^2} = \frac{1}{2}$.

%For the case of $l = 1$, the lowest energy level is at $n = 2$. Figure \ref{fig:l=1} demonstrates that our neural network accurately solves for the three lowest energy levels of this system.

%\begin{figure}[h]
%    \centering
%    \includegraphics[scale=.4]{l=1.pdf}
%    \caption{Top left diagram shows the model's eigenvalue weight during training. We find the the three lowest eigenvalues for this system, corresponding to $ n = 2,3,4$.The bottom left plot depicts the corresponding Schrodinger Equation loss. The column of plots on the right are the found eigenfunction solutions. }
%    \label{fig:l=1}
%\end{figure}

\begin{figure}[h]
    \centering
    \includegraphics[scale=.3]{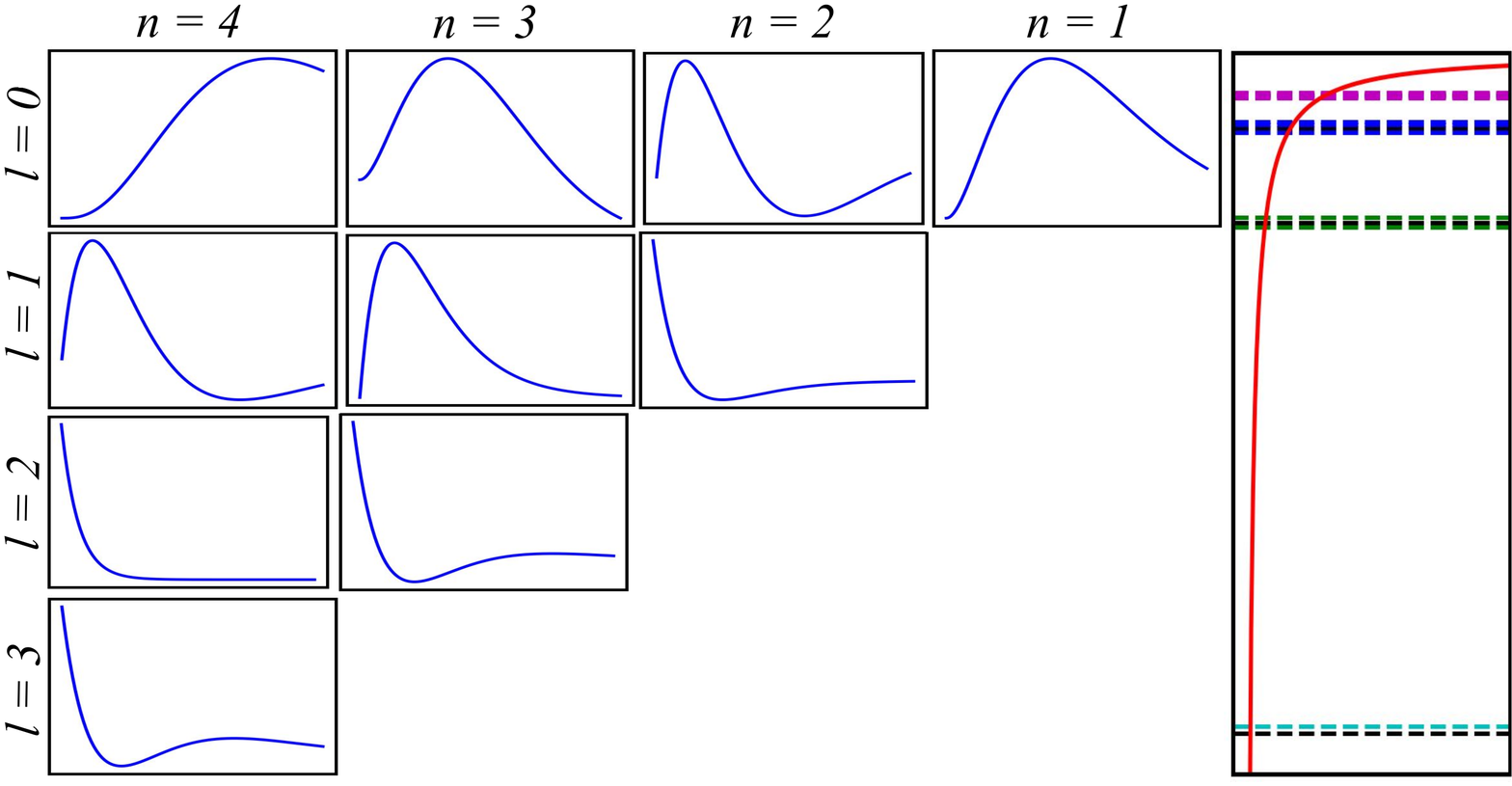}
    \caption{Upper triangular plots show the model's predicted eigenfunctions for angular momentum values $l = [0,3]$ inclusive, and for energy levels $n= [1,4]$ also inclusive. The rightmost plot shows the true eigenvalues in dashed black, with our method's found solutions in colors. The red line is the Coulomb potential function that describes the radial hydrogen atom problem.}
    \label{fig:hydrogen}
\end{figure}

We demonstrate that our method solves for the first few eigenfunctions for four different angular momentum values $l = 0, 1, 2, 3$. Figure \ref{fig:hydrogen} shows our model's solutions, arranged in a grid with angular momentum $l$ running along the vertical grid plots, and the energy level $n$ running along the horizontal axis.

Our method is able to solve for the lowest eigenvalue-eigenfunction pairs with good accuracy. We analysed the accuracy of our method's solutions in comparison to the true solutions which are analytically known. Table \ref{tab:performance} shows these results.

\begin{table}
\begin{tabular}{ |p{2.5cm}||p{2cm}|p{2cm}|  }
 \hline
 \multicolumn{3}{|c|}{Performance Results (\%)} \\
 \hline
 Schrodinger Problem & Eigenvalue Err. & Mean Squared Err. \\
 \hline
 Single ($n$ = 1, s)   & 0.00 & 8.9e-4 \\
 Single ($n$ = 1, a)   & 0.25 & 3.7e-4 \\
 Single ($n$ = 2, s)   & 0.25 & 9.1e-4 \\
 Single ($n$ = 2, a)   & 0.46 & 4.8e-4 \\
 \hline
 Double ($n$ = 1, s)   & 0.25 & 6.3e-4 \\
 Double ($n$ = 1, a)   & 0.32 & 4.8e-4 \\
 Double ($n$ = 2, s)   & 0.56 & 8.7e-4 \\
 Double ($n$ = 2, a)   & 0.61 & 7.1e-4 \\
 \hline
 H ($l$ = 0, $n$ = 1) & 1.46 & 4.0e-3 \\
 H ($l$ = 0, $n$ = 2) & 0.08 & 5.8e-3 \\
 H ($l$ = 0, $n$ = 3) & 0.70 & 8.2e-3 \\ %
 H ($l$ = 0, $n$ = 4) & 1.10 & 1.0e-2 \\ %
 H ($l$ = 1, $n$ = 2) & 4.08 & 3.8e-5 \\ %
 H ($l$ = 1, $n$ = 3) & 2.38 & 1.8e-4 \\ %
 H ($l$ = 1, $n$ = 4) & 2.52 & 2.2e-4 \\ %
 H ($l$ = 2, $n$ = 3) & 0.18 & 6.9e-6 \\
 H ($l$ = 2, $n$ = 4) & 0.48 & 6.2e-4 \\
 H ($l$ = 3, $n$ = 4) & 1.44 & 6.5e-3 \\
 \hline

\end{tabular}
\caption{This table presents comparisons with our model's solutions to the true analytical solutions. We find that our model finds the true eigenvalues to approximately 1 \% error consistently. Mean squared error denotes the mean squared error of the function divided by the maximum of the absolute value of the true eigenfunction.}
\label{tab:performance}
\end{table}

%%%%%%%%%%%%%%%%%%%%%%%%%%%%%%%%%%%%%%%%%%%%%%%%%%%%%%%%%%%%%%%%%%%%%%%%%%%%%%%%%%%%%%%%
\section{Conclusion}
In recent years, there has been a growing interest in the application of  neural networks to study differential equations. In this study, we introduced a neural network  that is capable  of discovering  eigenvalues and eigenfunctions for boundary conditioned differential eigenvalue problems. The obtained solutions identically satisfy the given boundary conditions via a parametric function. We imposed even and odd symmetry in the network structure  for problems that require such solutions, such as the single and multiple finite  wells. We also introduced an orthogonality loss, which allows the network to learn new eigenfunctions that are orthogonal to all previously learned eigenfunctions. Furthermore, a normalization loss was used to enforce that the learned solutions are not trivial solutions, and that the quantum physical interpretation of eigenfunctions as probability distributions can be supported. The optimization solely depends on the network's predictions, consisting of an unsupervised learning method. We  demonstrated the capability of the proposed architecture and training methodologies by solving  the finite  well, multiple finite  wells, and hydrogen atom quantum problems.

%%%%%%%%%%%%%%%%%%%%%%%%%%%%%%%%%%%%%%%%%%%%%%%%%%%%%%%%%%%%%%%%%%%%%%%%%%%%%%%%%%%%%%%%
\section{Future Research}
For future work, we will generalise our method in two ways. One generalisation is towards more dimensions. For instance, the full solutions to the hydrogen atom Schrodinger equation are three-dimensional. We believe such generalisations will also more clearly reveal the advantages of solving such equations with neural networks. It is also possible to extend into the temporal dimension and solve the time dependent Schrodinger equation. The other avenue for future research is to apply our method to more general eigenvalue differential equations. This paper focuses on the Schrodinger's equation, which belongs to the Sturm-Liouville family.  This study  lays the groundwork for using neural networks to solve any eigenvalue differential equation.

%%%%%%%%%%%%%%%%%%%%%%%%%%%%%%%%%%%%%%%%%%%%%%%%%%%%%%%%%%%%%%%%%%%%%%%%%%%%%%%%%%%%%%%%
\section{Broader Impact}
This work is valuable for computational physicists and applied mathematicians, as well as in any field where differential eigenvalue problems may arise. We have demonstrated our method's success for the one-dimensional Schrodinger equation, but the technique can be generalised to Sturm-Liouville problems, as well as higher dimensional equations (e.g. 3D Schrodinger and Helmholtz equations). We strongly believe that this study will serve as the groundwork for future work in the area of solving differential equations using deep learning methods.  We neither  foresee and nor desire our research results  to be used for any kind of discrimination.

%\section*{Acknowledgment}

%The authors would like to thank Dr. Georgios A. Tritsaris for the fruitful discussions. 

\bibliographystyle{IEEEtran}
\bibliography{refs}

\end{document}